\documentclass[10pt,twocolumn,letterpaper]{article}
\usepackage{cvpr}
\usepackage{times}
\usepackage{epsfig}
\usepackage{graphicx}
\usepackage{amsmath}
\usepackage{amssymb}
\usepackage{bm}
\usepackage{url}
\usepackage[normalem]{ulem}
 
\usepackage{enumitem}
\setlist{nolistsep}  

\usepackage[pagebackref=true,breaklinks=true,colorlinks,bookmarks=false]{hyperref}

\cvprfinalcopy 

\def\cvprPaperID{64} 

\begin{document}

\title{DeLiGAN : Generative Adversarial Networks for Diverse and Limited Data}

\author{Swaminathan Gurumurthy\thanks{Equal contribution}  \qquad Ravi Kiran Sarvadevabhatla\footnotemark[1] \qquad R. Venkatesh Babu \\ Video Analytics Lab, CDS, Indian Institute of Science\\
 Bangalore, INDIA 560012 \\ {\tt\small gauthamsindia95@gmail.com, ravika@gmail.com, venky@cds.iisc.ac.in}}

\maketitle

\begin{abstract}
     A class of recent approaches for generating images, called Generative Adversarial Networks (GAN), have been used to generate impressively realistic images of objects, bedrooms, handwritten digits and a variety of other image modalities. However, typical GAN-based approaches require large amounts of training data to capture the diversity across the image modality. In this paper, we propose DeLiGAN -- a novel GAN-based architecture for diverse and limited training data scenarios. In our approach, we reparameterize the latent generative space as a mixture model and learn the mixture model's parameters along with those of GAN. This seemingly simple modification to the GAN framework is surprisingly effective and results in models which enable diversity in generated samples although trained with limited data. In our work, we show that DeLiGAN can generate images of handwritten digits, objects and hand-drawn sketches, all using limited amounts of data. To quantitatively characterize intra-class diversity of generated samples, we also introduce a modified version of ``inception-score", a measure which has been found to correlate well with human assessment of generated samples.
\end{abstract}

\section{Introduction}
\label{sec:intro}

Generative models for images have enjoyed a resurgence in recent years, particularly with the availability of large datasets~\cite{russakovsky2015imagenet,zhou2014learning} and advent of deep neural networks~\cite{krizhevsky2012imagenet}. In particular, Generative Adversarial Networks (GANs)~\cite{goodfellow2014generative} and Variational Auto-Encoders (VAE)~\cite{kingma2013auto} have shown a lot of promise in this regard. In this paper, we focus on GAN-based approaches. 

A typical GAN framework consists of two components, a generator $G$ and a discriminator $D$. The generator $G$ is modelled so that it transforms a random vector $z$ into an image $I$, i.e. $I = G(z)$. $z$ usually arises from an easy-to-sample distribution (e.g. uniform). $G$ is trained to generate images $I$ which are indistinguishable from a sampling of the true distribution, i.e $I \sim p_{data}$, where $p_{data}$ is the true distribution of images. The discriminator $D$ takes an image as input and outputs the probability that the image is from the true data distribution $p_{data}$. In practice, $D$ is trained to output a low probability $p_D$ when fed a ``fake" (generated) image. $D$ and $G$ are trained adversarially to improve by competing with each other. A proper training regime ensures that at end of training, $G$ generates images which are essentially indistinguishable from real images, i.e. $p_D(G(z)) = 0.5$~\cite{goodfellow2014generative}. 

In recent times, GAN-based approaches have been used to generate impressively realistic house-numbers~\cite{chen2016infogan}, faces, bedrooms ~\cite{radford2015unsupervised} and a variety of other image categories~\cite{reed2016learning,salimans2016improved}. Usually, these image categories tend to have extremely complex underlying distributions. This complexity arises from two factors: (1) level of detail (e.g. color photos of objects have more detail than binary handwritten digit images) (2) diversity (e.g. inter and intra-category variability is larger for object categories compared to, say, house numbers). To be viable, generator $G$ needs to have sufficient capacity for tackling these complexity-inducing factors. Typically, such capacity is attained by having deep networks for $G$ ~\cite{BengioMDR13}. However, training high-capacity generators requires a large amount of training data. Therefore, existing GAN-based approaches are not viable when the amount of training data is limited.  

\textbf{Contributions:}

\begin{itemize}
    \item We propose DeLiGAN -- a novel GAN-based framework which is especially suited for small-yet-diverse data scenarios (Section \ref{sec:divaligan}). 
    \item We show that DeLiGAN enables generation of diverse images for a number of different modalities in limited data regimes. In particular, we construct modality-specific models which generate images of handwritten digits (Section \ref{sec:MNIST}), photo objects (Section \ref{sec:CIFAR10}) and hand-drawn sketches (Section \ref{sec:sketches}). 
    \item To quantitatively characterize the intra-class diversity of generated samples, we also design a modified version of the ``inception-score" ~\cite{salimans2016improved}, a measure which has been found to correlate well with human assessment of generated samples (Section \ref{sec:incpscore}). 
\end{itemize}

The rest of the paper is organised as follows: We give an overview of the related work in Section \ref{sec:relatedwork}, review GAN in Section \ref{sec:background} and then go on to describe our model DeLiGAN in Section \ref{sec:divaligan}. In Section \ref{sec:experiments}, we discuss experimental results which showcase the capabilities of our model. Towards the end of the paper, we discuss these results and the implications of our design decisions in Section \ref{sec:discussion}. We conclude with some pointers for future work in Section \ref{sec:conclusion}.

\section{Related Work}
\label{sec:relatedwork}

Generative Adversarial Networks (GANs) have recently gained a lot of popularity due to the relative sharpness of samples generated by these models compared to other approaches. The originally proposed baseline approach~\cite{goodfellow2014generative} has been modified to incorporate deep convolutional networks without destabilizing the training scheme and achieving significant qualitative improvements in image quality~\cite{denton2015deep,radford2015unsupervised}. Further improvements were made by Salimans et al.~\cite{salimans2016improved} by incorporating algorithmic tricks such as mini-batch discrimination which stabilize training and provide better image quality. We incorporate some of these tricks in our work as well.

Our central idea -- utilizing a mixture model for latent space -- has been suggested in various papers, but mostly in the context of variational inference. For example, Gershman et al.~\cite{gershman2012nonparametric}, Jordan et al.~\cite{jordan1999introduction} and Jaakkola et al.~\cite{jaakkola1998improving} model the approximate posterior of the inferred latent distribution as a mixture model to represent more complicated distributions. More recently, Renzede et al.~\cite{rezende2015variational} and Kingma et al.~\cite{kingma2016improving} propose `normalizing flows' to transform the latent probability density through a series of invertible mappings to construct a complex distribution. In the context of GANs, no such approaches exist, to the best of our knowledge. 

Our approach can be viewed as an attempt to modify the latent space to obtain samples in the high probability regions in the latent space. The notion of latent space modification has been explored in some recent works. For example, Han et al.~\cite{han2016alternating} propose to alternate between training the latent factors and the generator parameters. Arulkumaran et al.~\cite{arulkumaran2016improving} formulate an MCMC sampling process to sample from high probability regions of a learned latent space in variational or adversarial autoencoders.

\begin{figure*}[!t]
\centering
\includegraphics[width=0.75\textwidth]{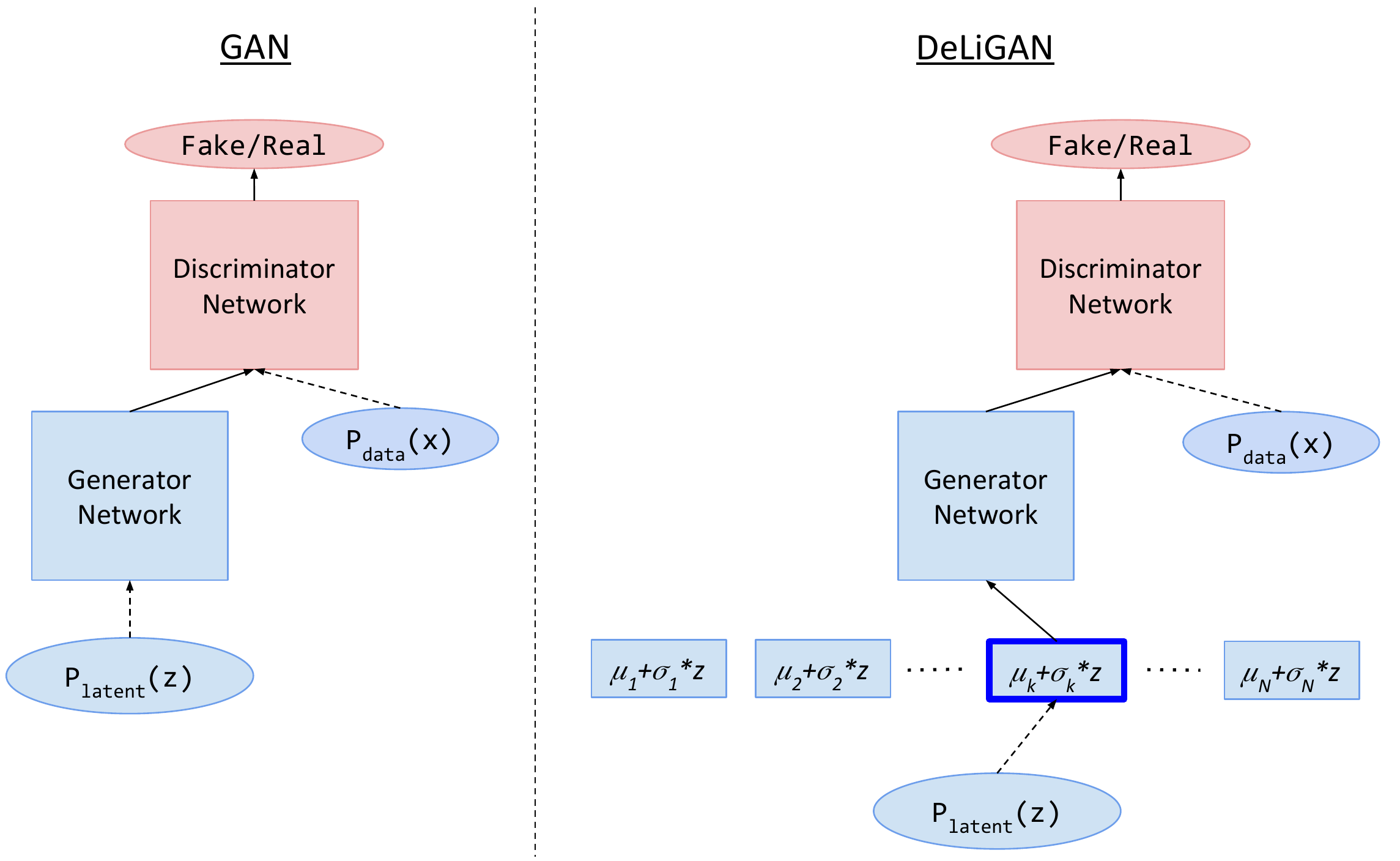}
\caption{Baseline GAN framework (left of dotted line) and our model - DeLiGAN (right of dotted line). Dotted arrows indicate sampling from the source. Instead of sampling directly from a simple latent distribution as done for baseline model, we reparameterize the latent space using a mixture of Gaussian model in DeLiGAN. We randomly select one of the Gaussian components (depicted with a dark blue outline in the right-side figure) and employ the ``reparameterization trick"~\cite{kingma2013auto} to obtain a sample from the chosen Gaussian. See Section \ref{sec:divaligan} for details.}
\label{fig:gan-DeLiGAN}
\end{figure*}

\section{Generative Adversarial Networks (GANs)}
\label{sec:background}

Although GANs were introduced in Section \ref{sec:intro}, we formally describe them below to establish continuity.

A typical GAN framework consists of two components, a generator $G$ and a discriminator $D$. In practice, these two components are usually two neural networks. The generator $G$ is modelled so that it transforms a random vector $z$ into an image $x_{G}$, i.e. $x_{G} = G(z)$. $z$ typically arises from an easy-to-sample distribution, for e.g. $z \sim \mathcal{U}(-1,1)$ where $\mathcal{U}$ denotes a uniform distribution. $G$ is trained to generate images which are indistinguishable from a sampling of the true distribution. In other words, while training $G$, we try to maximise $p_{data}(x_G)$, the probability that the generated samples belong to the data distribution.

\begin{eqnarray}p_{data}(x_{G})  &=& \int_z p(x_{G},z) dz \\
&=& \int_z p_{data}(x_{G}|z) p_{z}(z) dz
\label{eqn:gendatalkhood}
\end{eqnarray}

The above equations make explicit the fact that GANs assume a fixed, easy to sample, prior distribution $p_{z}(z)$ and then maximize $p_{data}(x_{G}|z)$ by training the generator network to produce samples from the data distribution. 

The discriminator $D$ takes an image $I$ as input and outputs the probability $p_D(I)$ that the image is from the true data distribution. Typically, $D$ is trained to output a low probability when fed a ``fake" (generated) image. Thus, $D$ is supposed to act as an expert, estimating the probability that the sample is from the true data distribution as opposed to the $G$'s output.

$D$ and $G$ are trained adversarially to improve by competing with each other.  This is achieved by alternating between the training phases of $D$ and $G$. $G$ tries to `fool' $D$ into thinking that its outputs are from the the true data distribution by maximizing its score $D(G(z))$. This is achieved by solving the following optimization problem in the generator phase of training:

\begin{equation}
\min_{G} V_{G} (D, G) = \displaystyle \min_{G} \Big( \mathbb{E}_{z \sim p_{z}}[log(1-D(G(z)))] \Big)
\label{eqn:gloss}
\end{equation}

On the other hand, $D$ tries to minimize the score it assigns to generated samples $G(z)$ by minimising $D(G(z))$ and maximize the score it assigns to the real (training) data $x$ by maximising $D(x)$. Hence, the optimisation problem for $D$ can be formulated as follows:

\begin{equation}
\begin{split}
\max_{D} V_{D} (D, G) = & \max_{D} \Big( \mathbb{E}_{x \sim p_{data}}[log D(x)] \\
& + \mathbb{E}_{z \sim p_{z}}[log(1-D(G(z)))] \Big)
\end{split}
\label{eqn:dloss}
\end{equation}

Hence the combined loss for the GAN can now be written as:

\begin{equation}
\begin{split}
\min_{G}\max_{D} V (D, G) = & \min_{G}\max_{D} \Big( \mathbb{E}_{x \sim p_{data}}[log D(x)] \\
& + \mathbb{E}_{z \sim p_{z}}[log(1-D(G(z)))] \Big)
\end{split}
\label{eqn:ganoptimizationeq}
\end{equation}

In their work, Goodfellow et al.~\cite{goodfellow2014generative} show that Equation \ref{eqn:ganoptimizationeq} gives us Jensen--Shannon (JS) divergence between the model's distribution and data generating process. A proper training regime ensures that at the end of training, $G$ generates images which are essentially indistinguishable from real images, i.e. $p_D(G(z)) = 0.5$ and JS divergence achieves its lowest value.

\section{Our model - DeLiGAN}
\label{sec:divaligan}

\begin{figure*}[!ht]
\centering
\includegraphics[width=\textwidth]{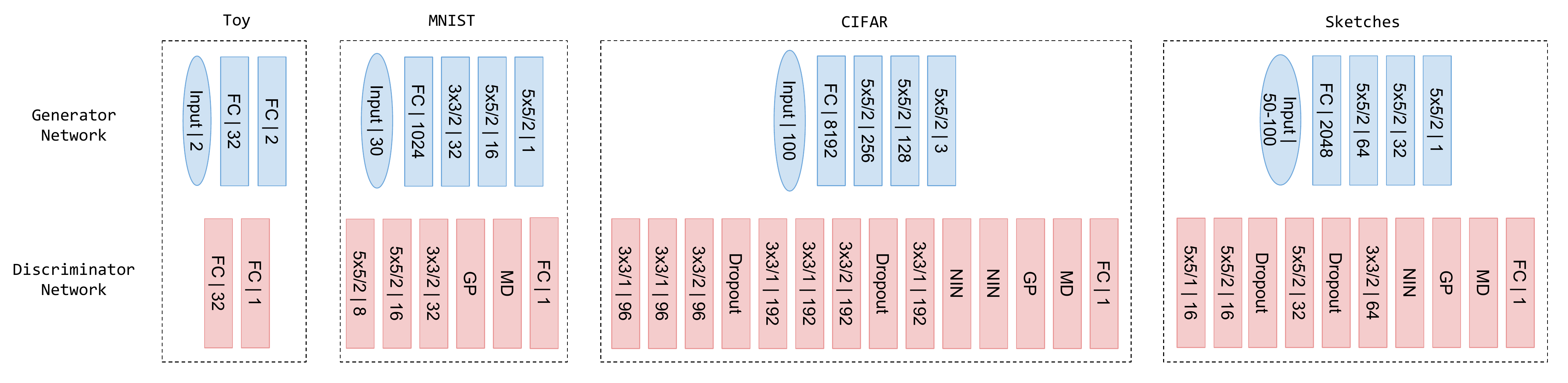}
\caption{Architectural details of Generator and Discriminator in GAN models experimentally evaluated for various image domains. Notation: FC=Fully Connected Layer, GP = Global Pooling, NIN = Network-in-Network, MD=mini-batch discrimination. Convolutional layers are specified in the format dimensions/\hspace{0.5mm}stride\hspace{0.5mm} $\vert$ \hspace{0.5mm}number of filters.}
\label{fig:architectures}
\end{figure*}

In GAN training, we essentially attempt to learn a mapping from a simple latent distribution $p_{z}$ to the complicated data distribution (Equation \ref{eqn:gendatalkhood}). This mapping requires a deep generative network which can disentangle the underlying factors of variation in the data distribution and enable diversity in generated samples ~\cite{BengioMDR13}. In turn, this translates to the requirement of large amounts of data. Therefore, when data is limited yet originates from a  diverse image modality, increasing the network depth becomes infeasible. Our solution to this conundrum is the following: Instead of increasing the model depth, we propose to increase the modelling power of the prior distribution. In particular, we propose a reparameterization of the latent space as a Mixture-of-Gaussians model (see Figure \ref{fig:gan-DeLiGAN}). 

\begin{equation}
    p_{z}(z) = \sum_{i=1}^{N} \phi_{i} g(z|\mu_{i},\Sigma_{i})
\end{equation}

where $g(z|\mu_{i},\Sigma_{i})$ represents the probability of the sample $z$ in the normal distribution, $\mathcal{N}(\mu_{i},\Sigma_{i})$. 
For reasons which will be apparent shortly (Section \ref{sec:training}), we assume uniform mixture weights, $\phi_{i}$, i.e. 

\begin{equation}
	p_{z}(z) = \sum_{i=1}^{N} \frac{g(z|\mu_{i},\Sigma_{i})}{N} 
\label{eqn:zmog}
\end{equation}

To obtain a sample from the above distribution, we randomly select one of the $N$ Gaussian components and employ the ``reparameterization trick" introduced by Kingma et al.~\cite{kingma2013auto} to sample from the chosen Gaussian. We also assume that each Gaussian component has a diagonal covariance matrix. Suppose the $i$-th Gaussian is chosen.  Let us denote the diagonal elements of the corresponding covariance matrix as $\sigma_i = [\sigma_{i_1} \sigma_{i_2} \ldots \sigma_{i_K}]$ where $K$ is the dimension of the latent space. For the ``reparameterization trick", we represent the sample from the chosen $i$-th Gaussian as a deterministic function of $\mu_{i}$,  $\sigma_{i}$ and an auxiliary noise variable $\epsilon$.

\begin{equation}
z = \mu_{i} + \sigma_{i} \epsilon  \text{ where $\epsilon \sim \mathcal{N}(0,1)$}
\label{eqn:reparamtrick}
\end{equation}

Therefore, obtaining a latent space sample translates to sampling $\epsilon \sim \mathcal{N}(0,1)$ and calculating $z$ according to Equation \ref{eqn:reparamtrick}. Substituting Equations \ref{eqn:zmog}, \ref{eqn:reparamtrick} in RHS of Equation \ref{eqn:gendatalkhood}, we get: 

\begin{equation}
	p_{data}(G(z)) =  \sum_{i=1}^{N} \int \frac { p_{data}( G(\mu_{i}+\sigma_{i}\epsilon) | \epsilon) \hspace{0.2mm} p(\epsilon) d\epsilon}{N}
\label{eqn:pdatagz}
 \end{equation}
 
Let us define $\bm\mu = [\mu_1 , \mu_2 , \ldots , \mu_N]^T$ and $\bm\sigma = [\sigma_1 , \sigma_2 ,\ldots ,\sigma_N]^T$. Therefore, our new objective is to learn $\bm\mu$ and $\bm\sigma$ (along with the GAN parameters) to maximise $p_{data}(G(\mu_{i}+\sigma_{i}\epsilon)|\epsilon)$. 

Next, we describe the procedure for learning $\bm\mu$ and $\bm\sigma$. 

\begin{figure*}[!ht]
\centering
\includegraphics[width=\textwidth]{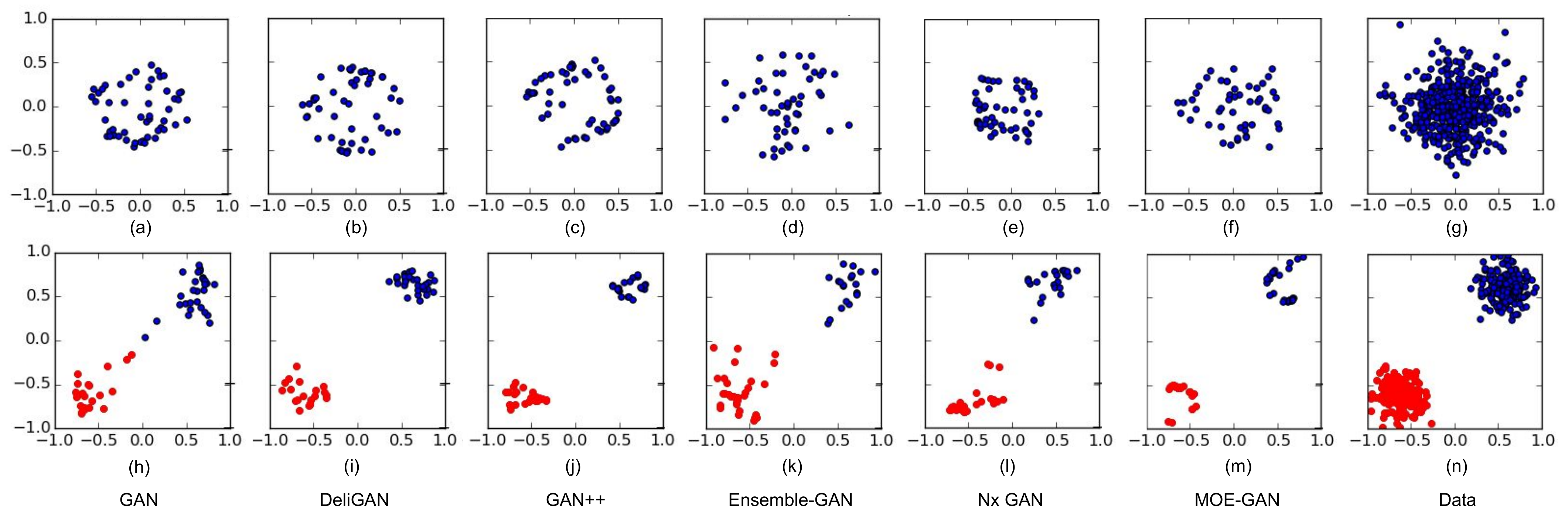}
\caption{Comparing the performance of baseline GANs and our model (DeLiGAN) for toy data. Refer to Section \ref{sec:toydata} for details.}
\label{fig:toydata}
\end{figure*}

\subsection{Learning $\bm\mu$ and $\bm\sigma$}
\label{sec:training}

For each Gaussian component,  we first need to initialise its parameters.  For $\bm \mu_{i}, 1 \leqslant i \leqslant N$, we sample from a simple prior -- in our case, a uniform distribution $\mathcal{U}(-1,1)$. For $\sigma_i$, we assign a small, fixed non-zero initial value ($0.2$ in our case). Normally, the number of samples we generate from each Gaussian relative to the other Gaussians during training gives us a measure of the `weight'  $\bm\pi$ for that component. However, $\bm\pi$ is not a trainable parameter in our model since we cannot obtain gradients for $\pi_{i}$s. Therefore, as mentioned before, we consider all components to be equally important. 

To generate data, we randomly choose one of the $N$ Gaussian components and  sample a latent vector $z$ from the chosen Gaussian (Equation \ref{eqn:reparamtrick}). $z$ is passed to $G$ to obtain the output data (image). The generated sample $z$ can now be used to train parameters of $D$ or $G$ using the standard GAN training procedure (Equation \ref{eqn:ganoptimizationeq}). In addition, $\bm\mu$ and $\bm\sigma$ are also trained simultaneously along with $G$'s parameters, using gradients arising from $G$'s loss function. 

However, we need to consider a subtle issue here involving ${\bm\sigma}$. Since $p_{data}(G(z))$ (Equation \ref{eqn:pdatagz}) has local maxima at the $\mu_{i}$s, $G$ tries to decrease the $\sigma_{i}$s in an effort to obtain more samples from the high probability regions. As a result $\sigma_i$s can collapse to zero. Hence, we add a ${L_2}$ regularizer to the generator cost to prevent this from happening. The original formulation of loss function for $G$ (Equation \ref{eqn:gloss}) now becomes: 

\begin{equation}
\begin{split}
\min_{G} V_G (D,G) = \min_{G} \mathbb{E}_{z \sim p_{z}}[log(1-D(G(z)))] \\+ \lambda \sum_{i=1}^{N}\frac{(1-\sigma_{i})^2}{N}
\end{split}
\end{equation}

Note that this procedure can be extended to generate a batch of images for mini-batch training. Indeed, increasing the number of samples per Gaussian increases the accuracy of the gradients used to update $\bm\mu$ and $\bm\sigma$ since they are averaged out over $p(\epsilon)$~\cite{burda2015importance}, thereby speeding up training. 

\section{Experiments}
\label{sec:experiments}

For our DeLiGAN framework, the choice of $N$, the number of Gaussian components, is made empirically -- more complicated data distributions require more Gaussians. Larger values of $N$ potentially help model with relatively increased diversity. However, increasing $N$ also increases memory requirements. Our experiments indicate that increasing $N$ beyond a point has little to no effect on the model capacity since the Gaussian components tend to `crowd' and become redundant. We use a $N$ between $50$ and $100$ for our experiments. 

To quantitatively characterize the diversity of generated samples, we also design a modified version of the ``inception-score", a measure which has been found to correlate well with human evaluation ~\cite{salimans2016improved}. We describe this score next.

\subsection{Modified Inception Score}
\label{sec:incpscore}

Passing a generated image $x=G(z)$ through a trained classifier with an ``inception" architecture~\cite{szegedy2015going} results in a conditional label distribution $p(y|x)$. If $x$ is realistic enough, it should result in a ``peaky" label distribution i.e. $p(y|x)$ should have low entropy. We also want all categories to be covered uniformly among the generated samples, i.e. $ p(y) = \int_z p(y|x = G(z)) p_{z}(z)dz$ should have high entropy. These two requirements are unified into a single measure called ``inception-score" as  $e^{\mathbb{E}_x KL( p(y|x) || p(y) )}$ where $KL$ stands for KL-divergence and expectation $\mathbb{E}$ is taken over generated samples $x$. 

\textbf{Our modification:} In its original formulation, ``inception-score" assigns a higher score for models that result in a low entropy class conditional distribution $p(y|x)$. However, it is desirable to have diversity within image samples of a particular category. To characterize this diversity, we use a cross-entropy style score $-p(y|x_{i})log(p(y|x_{j}))$ where $x_{j}$s are samples of the same class as $x_{i}$ as per the outputs of the trained inception model. We incorporate this cross-entropy style term into the original ``inception-score" formulation and define the modified ``inception-score" (m-IS) as a KL-divergence: $e^{\mathbb{E}_{x_{i}}[\mathbb{E}_{x_{j}}[(\mathbb {KL}(P(y|x_{i})||P(y|x_{j})) ]]}$. Essentially, m-IS can be viewed as a proxy for measuring intra-class sample diversity along with the sample quality. In our experiments, we report m-IS scores on a per-class basis and a combined m-IS score averaged over all classes.

We analyze the performance of DeLiGAN models trained on toy data, handwritten digits~\cite{mnistlecun}, photo objects~\cite{krizhevsky2009learning} and hand-drawn object sketches~\cite{eitz2012humans} and compare with a regular GAN model. Specifically, we use a variant of DCGAN~\cite{radford2015unsupervised} with mini-batch discrimination in the discriminator~\cite{salimans2016improved}. We also need to note here that DeLiGAN adds extra parameters over DCGAN. Therefore, we also compare DeLiGAN with baseline models containing an increased number of learnable parameters. We start by describing a series of experiments on toy data. 

\subsection{Toy Data}
\label{sec:toydata}

As a baseline GAN model for toy data, we set up a multi-layer perceptron with one hidden layer as $G$ and $D$ (see Figure \ref{fig:architectures}). For the DeLiGAN model, we incorporate the mixture of Gaussian layer as shown in Figure \ref{fig:gan-DeLiGAN}. We also compare DeLiGAN with four other baseline models -- (i) GAN++ (instead of mixture of Gaussian layer, we add a fully connected layer containing $N$ neurons between the input ($z$) and the generator) (ii) Ensemble-GAN (An ensemble-of-$N$-generators setting for DeLiGAN. During training, we randomly choose one of the generators $G_i$ for training and update its parameters along with $\mu_i,\sigma_i$) (iii) $N$x-GAN (We increase number of parameters in the generator network $N$ times by having $N$ times more neurons in the hidden layer) and (iv) MoE-GAN (This is short for Mixture-of-Experts GAN. In this model, we just append a uniform discrete variable via a $N$-dimensional one-hot encoding~\cite{chen2016infogan} to the random input $z$). 

For the first set of experiments, we design our generator network to output data samples originally belonging to a unimodal 2-D Gaussian data (see Figure \ref{fig:toydata}(g)). Figures \ref{fig:toydata} (a)-(f) show samples generated by the respective GAN variants for this data. For the unimodal case, all models perform reasonably well in generating samples. 

For the second set of experiments, we replace the unimodal distribution with a bi-modal distribution comprising two Gaussians (Figure \ref{fig:toydata}(n)). The results in this case show that DeLiGAN is able to clearly model the two separate distributions whereas the baseline GAN frameworks struggle to model the void in between (Figure \ref{fig:toydata}(h-m)). Although the other variants, containing more parameters, were able to model the two modes, they still struggle to model the local structure in the Gaussians properly. The generations produced by DeLiGAN look the most convincing. Although not obvious from the results, a recurring trend across all the baseline models was the relative difficulty in training due to instabilities. On the other hand, training DeLiGAN was much easier in practice. As we shall soon see, this phenomenon of suboptimal baseline models and better performance by DeLiGAN persists even for more complex data distributions (CIFAR-10, sketches etc.)

\begin{figure}[!ht]
\centering
\includegraphics[width=0.5\textwidth]{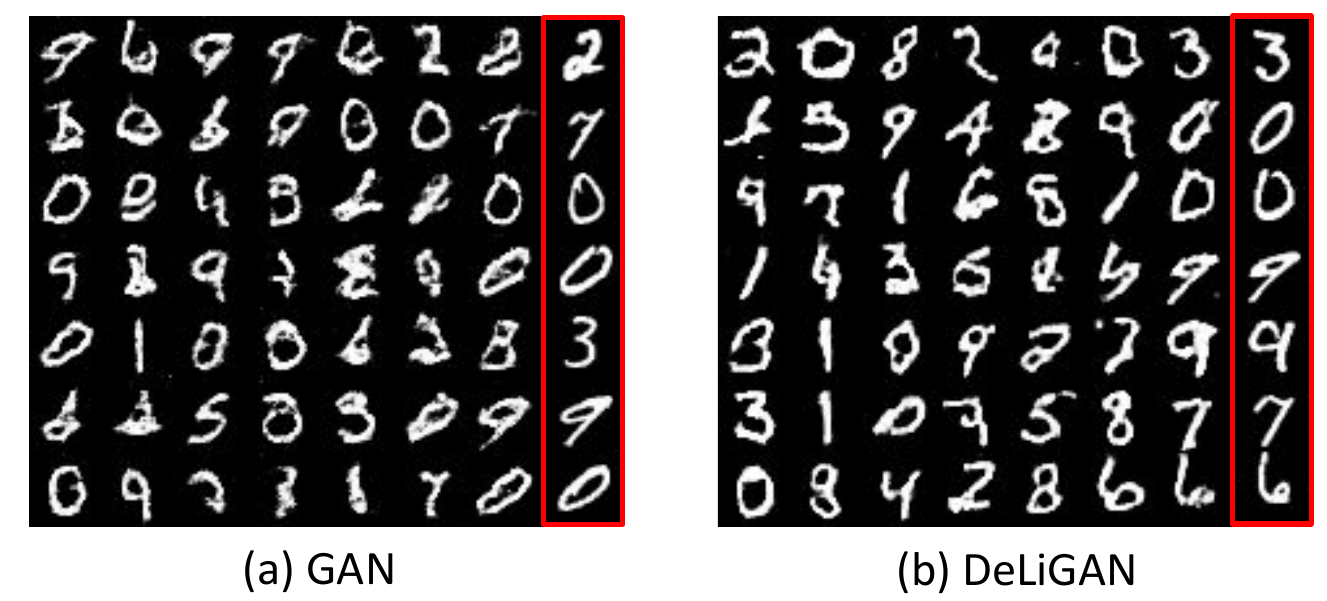}
\caption{Comparing the performance of GAN and our model (DeLiGAN) for MNIST handwritten digits data. Refer to Section \ref{sec:MNIST} for details.}
\label{fig:MNIST}
\end{figure}

\begin{table*}[!t]
\begin{center}
\resizebox{\textwidth}{!}{%
\begin{tabular}{|c|c|c|c|c|c|c|c|c|c|c||c|} 
\hline
& Plane & Car & Bird & Cat &  Deer & Dog & Frog & Horse & Ship & Truck & Overall \\
\hline 
GAN  &  $2.72 \pm 0.20$ & $2.02 \pm 0.18$ & $2.21 \pm 0.44$ & $\mathbf{2.43} \pm 0.19$ & $2.06 \pm 0.09$ & $\mathbf{2.22} \pm 0.23$ & $1.82 \pm 0.08$ & $2.12 \pm 0.55$ & $1.19 \pm 0.19$ & $\mathbf{2.16} \pm 0.15$ & $2.15 \pm 0.25$ \\
\hline
DeLiGAN & $ \mathbf{2.78} \pm 0.02$ & $ \mathbf{2.36} \pm 0.06$ & $ \mathbf{2.44} \pm 0.07$ & $2.17 \pm 0.04$ & $ \mathbf{2.31} \pm 0.02$ & $1.27 \pm 0.01$ & $\mathbf{2.31} \pm 0.02$ & $\mathbf{3.63} \pm 0.14$ & $\mathbf{1.51} \pm 0.03$ & $2.00 \pm 0.05$ & $\mathbf{2.28} \pm 0.62$ \\
\hline 
MoE-GAN & $2.69 \pm 0.08$ & $2.08 \pm 0.05$ & $2.01 \pm 0.06$ & $2.19 \pm 0.04$ & $2.16 \pm 0.03$ & $1.85 \pm 0.09$ & $1.84 \pm 0.07$ & $2.14 \pm 0.08$ & $1.60 \pm 0.04$ & $1.85 \pm 0.05$ & $2.04 \pm 0.28$ \\
\hline 
GAN++ & $2.44 \pm 0.06$ & $1.73 \pm 0.04$ & $1.68 \pm 0.05$ & $2.27 \pm 0.06$ & $2.23 \pm 0.04$ & $1.73 \pm 0.03$ & $1.56 \pm 0.02$ & $1.21 \pm 0.04$ & $1.25 \pm 0.02$ & $1.53 \pm 0.02$ & $1.76 \pm 0.40$ \\
\hline 
\end{tabular}
}
\end{center}
\caption{Comparing modified ``inception-score" values for baseline GANs and DeLiGAN across the $10$ categories of CIFAR-10 dataset. Larger scores are better. The entries represent score's mean value and standard deviation for the category.}
\label{tab:mis-cifar}
\end{table*}

\subsection{MNIST}
\label{sec:MNIST}

The MNIST dataset contains $60,000$ images of handwritten digits from $0$ to $9$~\cite{mnistlecun}. We conduct experiments on a reduced training set of $500$ images to mimic the low-data scenario. The images are sampled randomly from the dataset, keeping the total number of images per digit constant. For MNIST, the generator network has a fully connected layer followed by $3$ deconvolution layers while the discriminator network has $3$ convolutional layers followed by a mini-batch discrimination layer (see Figure \ref{fig:architectures}). 

In Figure \ref{fig:MNIST}, we show typical samples generated by both models, arranged in a $7 \times 7$ grid. For each model, the last column of digits (outlined in red), contains nearest-neighbor images (from the training set) to the samples present in the last ($7$th) column of the grid. For nearest neighborhood computation, we use $L_2$ distance between the images. 

The samples produced by our model (Figure \ref{fig:MNIST}(b), right) are visibly crisper compared to baseline GAN (Figure \ref{fig:MNIST}(a), left). Also, some of the samples produced by the GAN model are almost identical to one other (shown as similarly colored boxes in Figure \ref{fig:MNIST}(a)) whereas our model produces more diverse samples. We also observe that some of the samples produced by the baseline GAN model are deformed and don't resemble any digit. This artifact is much less common in our model. Additionally, in practice, the baseline GAN model frequently diverges during training given the small data regime and the deformation artifact mentioned above becomes predominant, eventually leading to homogeneous non-digit like samples. In contrast, our model remains stable during training and generates samples with better diversity.


\begin{figure}[!t]
\centering
\includegraphics[width=0.5\textwidth]{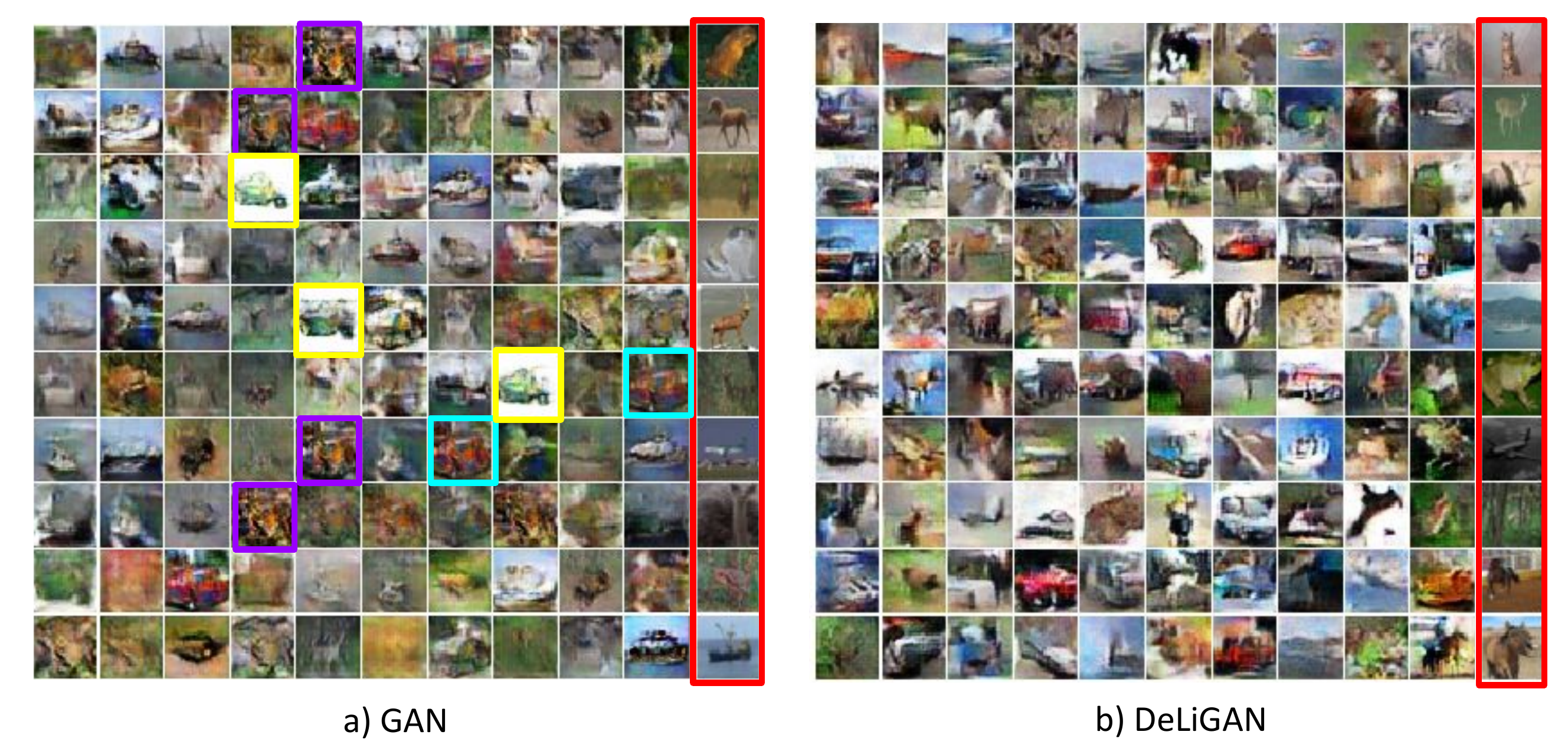}
\caption{Comparing the performance of GAN and our model (DeLiGAN) for CIFAR-10 data. Refer to Section \ref{sec:CIFAR10} for details.}
\label{fig:CIFAR10}
\end{figure}
\subsection{CIFAR 10}
\label{sec:CIFAR10}

The CIFAR 10 dataset~\cite{krizhevsky2009learning} contains $60,000$ $32 \times 32$ color images across $10$ object classes. Once again, to mimic the diverse-yet-limited-data scenario, we compare the architectures on a reduced dataset of $2000$ images. The images are drawn randomly from the entire dataset, keeping the number of images per category constant. For the experiments involving CIFAR dataset, we adopt the architecture proposed by Goodfellow \etal~\cite{goodfellow2014generative}. The generator has a fully connected layer followed by $3$ deconvolution layers with batch normalisation after each layer. The discriminator network has $9$ convolutional layers with dropout and weight normalisation, followed by a mini-batch discrimination layer.

Figure \ref{fig:CIFAR10} shows samples generated by our model and the baseline GAN model. As in the case of MNIST, some of the samples generated by the GAN, shown with similar colored bounding boxes, look nearly identical (Figure \ref{fig:CIFAR10}(a)). Again, we observe that our model produces visibly diverse looking samples and provides more stability. The modified ``inception-score" values for the models (Table \ref{tab:mis-cifar}) attest to this observation as well. Note that there exist categories (`cat', `dog') with somewhat better diversity scores for GAN. Since images belonging to these  categories are similar, these kinds of images would be better represented in the data. As a result, GAN performs better for these categories, whereas DeLiGAN manages to capture even the other under-represented categories. Table \ref{tab:mis-cifar} also shows the modified inception scores for the GAN++ and MoE-GAN models introduced in the toy experiments. We observe that the performance in this case is actually worse than the baseline GAN model, despite the increased number of parameters. Moreover, adding fully connected layers in the generator in GAN++ also leads to increased instability in training. We  hypothesize that the added set of extra parameters worsens the performance given our limited data scenario. In fact, for baseline models such as Ensemble-GAN and $N$x-GAN, the added set of parameters also makes computations prohibitively expensive.

Overall, the CIFAR dataset experiments demonstrate that our model can scale to more complicated real life datasets and still outperform the traditional GANs in low data scenarios.

\subsection{Freehand Sketches}
\label{sec:sketches}

\begin{figure*}[!ht]
\centering
\includegraphics[width=1.0\textwidth]{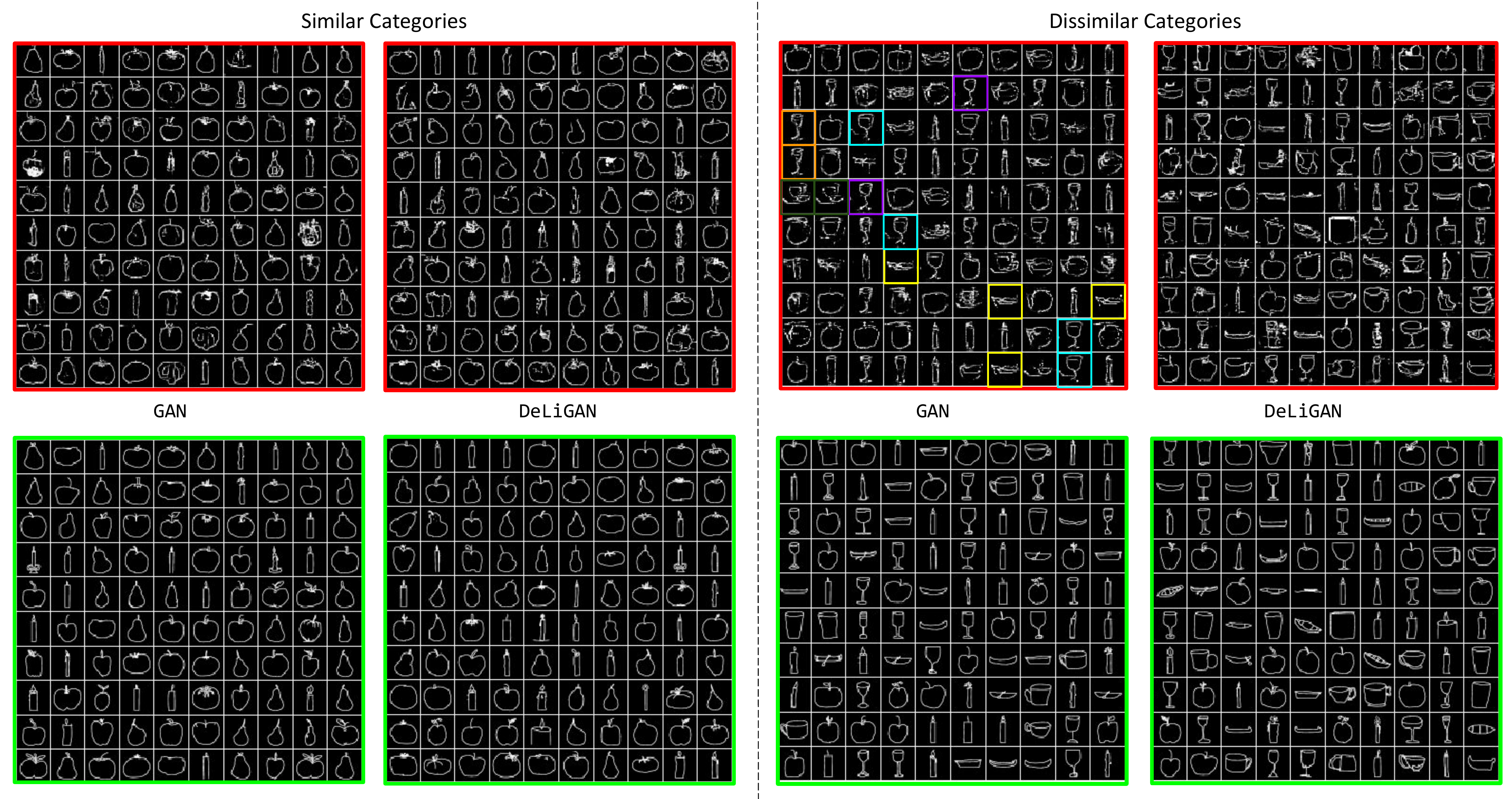}
\caption{Comparing the performance of GAN and our model (DeLiGAN) for hand-drawn sketches for similar categories (left side of dotted line) and dissimilar categories (right side of dotted line). Panels outlined in red correspond to generated samples. Panels outlined in green correspond to the nearest training examples. Similarly colored boxes for GAN generations of dissimilar categories indicate samples which look `similar'. Refer to Section \ref{sec:sketches} for details.}
\label{fig:sketches}
\end{figure*}

The TU-Berlin dataset~\cite{eitz2012humans}, contains $20,000$ hand-drawn sketches evenly distributed among $250$ object categories, which amounts to $80$ images per category. This dataset represents a scenario where the amount of training data is actually limited, unlike previous experiments where the quantity of training data was artificially restricted. For sketches, our network contains $5$ convolutional layers in the discriminator with weight normalization and dropout followed by mini-batch discrimination and $3$ deconvolutional layers, followed by a fully connected layer in the generator. To demonstrate the capability of our model, we perform two sets of experiments. 

For the first set of experiments, we select $4$ sketch categories -- \texttt{apple, pear, tomato, candle}. These categories have simple global contours, low sketch stroke density and are somewhat similar in appearance. During training, we augment the dataset using the flipped versions of the images. Once again, we compare the generated results of GAN and DeLiGAN.  Figure \ref{fig:sketches} shows the samples generated by DeLiGAN and GAN respectively, trained on the similar looking categories (left side of the dotted line). The samples generated by both the models look visually appealing. Our guess is that since the object categories are very similar, the data distribution can be easily modelled as a continuous distribution in the latent space. Therefore, the latent space doesn't need a multi-modal representation in this case. This is also borne out by the m-IS diversity scores in Table \ref{tab:missketchsimilar}.

\begin{table}[!t]
\begin{center}
\resizebox{0.98\linewidth}{!}{%
\begin{tabular}{|c|c|c|c|c||c|} 
\hline
 & Apple & Tomato & Pear & Candle & Overall \\
\hline 
GAN  & $1.31 \pm 0.01$ & $1.39 \pm 0.01$ & $\mathbf{1.49} \pm 0.03$ & $\mathbf{1.25} \pm 0.01$ & $\mathbf{1.36} \pm 0.09$ \\
\hline
DeLiGAN & $\mathbf{1.40} \pm 0.00$ & $\mathbf{1.9} \pm 0.00$ & $1.47 \pm 0.01$ & $1.22 \pm 0.01$  & $1.35 \pm 0.10$ \\
\hline 
\end{tabular}
}
\end{center}
\caption{Comparing modified ``inception-score" values for GAN and DeLiGAN across sketches from the $4$ `similar' categories. The entries represent score's mean value and standard deviation for the category.}
\label{tab:missketchsimilar}
\end{table}

For the second set of experiments, we select $5$ diverse looking categories -- \texttt{apple, wine glass, candle, canoe, cup} -- and compare the generation results for both the models. The corresponding samples are shown in Figure \ref{fig:sketches} (on the right side of the dotted line). In this case, DeLiGAN samples are visibly better, less hazy, and arise from a more stable training procedure. The samples generated by DeLiGAN also exhibit larger diversity, visibly and according to m-IS scores as well (Table \ref{tab:missketchdissimilar}).

\section{Discussion}
\label{sec:discussion}

The experiments described above demonstrate the benefits of modelling the latent space as a mixture of learnable Gaussians instead of the conventional unit Gaussian/uniform distribution. One reason for our performance is derived from the fact that mixture models can approximate arbitrarily complex latent distributions, given a sufficiently large number of Gaussian components. 

In practice, we also notice that our mixture model approach also helps increase the model stability and is especially useful for diverse, low-data regimes where the latent distribution might not be continuous. Consider the following: The gradients on $\mu_{i}$s push them in the latent space in a direction which increases the discriminator score, $D(G(z))$ as per the gradient update (Equation \ref{eqn:mugrad2}). Thus, samples generated from the updated Gaussian components result in higher probability, $p_{data}(G(z))$.

 \begin{equation}
	\frac{\partial V}{\partial \mu} = - \frac{1}{1-D(G(z))}\frac{\partial D(G(z))}{\partial G(z)}\frac{\partial G(z)}{\partial z}*1
\label{eqn:mugrad2}
 \end{equation}

 
Hence, as training progresses, we find the $\mu_{i}$s in particular, even if initialised in the lower probability regions, slowly drift towards the regions that lead to samples of high probability, $p_{data}(G(z))$. Hence, fewer points are sampled from the low probability regions. 
This is illustrated by (i) the locations of samples generated by our model in the toy experiments (Figure \ref{fig:toydata}(d)) (ii) relatively small frequency of bad quality generations (that don't resemble any digit) for the MNIST experiments (Figure \ref{fig:MNIST}). Our model successfully handles the low probability void between the two modes in the data distribution by emulating the void into its own latent distribution. As a result, no samples are produced in these regions. This can also be seen in the MNIST experiments -- our model produces very few non-digit like samples compared to the baseline GAN (Figure \ref{fig:MNIST}).

\begin{table}[!t]
\begin{center}
\resizebox{0.98\linewidth}{!}{%
\begin{tabular}{|c|c|c|c|c|c||c|} 
\hline
 & Wineglass & Candle & Apple & Canoe & Cup & Overall \\
\hline 
GAN  & $1.80 \pm 0.01$ & $1.48 \pm 0.02$ & $1.50 \pm 0.01$ & $1.53 \pm 0.01$ & $1.74 \pm 0.01$ & $1.61 \pm 0.13$ \\
\hline
DeLiGAN & $\mathbf{2.09} \pm 0.01$ & $\mathbf{1.57} \pm 0.02$ & $\mathbf{1.65} \pm 0.01$ & $\mathbf{1.75} \pm 0.01$ & $\mathbf{1.87} \pm  0.02$ & $\mathbf{1.79} \pm 0.18$ \\
\hline 
\end{tabular}
}
\end{center}
\caption{Comparing modified ``inception-score" values for GAN and DeLiGAN across sketches from the $5$ `dissimilar' categories. The entries represent score's mean value and standard deviation for the category.}
\label{tab:missketchdissimilar}
\end{table}

In complicated multi-modal settings, the data may be disproportionally distributed among the modes such that some of the modes contain relatively more data points. In this situation, the generator in baseline GAN tends to fit the latent distribution to the mode with maximum data as dictated by the Jensen-Shannon Divergence~\cite{theis2015note}. This results in low diversity among the generated samples since a section of the data distribution is sometimes overlooked by the generator network. This effect is especially pronounced in low data regimes because the number of modes in the image space increase due to the non-availability of data connecting some of the modes. As a result, the generator tries to fit to a small fraction of the already limited data. This is consistent with our experimental results wherein the diversity and quality of samples produced by baseline GANs deteriorate with decreasing amounts of training data (MNIST -- Figure \ref{fig:MNIST}, CIFAR  -- Figure \ref{fig:CIFAR10}) or increasing diversity of the training data (Sketches -- Figure \ref{fig:sketches}).  

Our design decision of having a trainable mixture model for latent space can be viewed as an algorithmic ``plug-in" that can be added to almost any GAN framework including recently proposed models~\cite{zhao2016energy,salimans2016improved} to obtain better performance on diverse data. Finally, it is also important to note that our model is still constrained by the modelling capacity of the underlying GAN framework itself. Hence, as we employ better GAN frameworks on top of our mixture of Gaussians layer, we can expect the model to generate realistic, high-quality samples.

\section{Conclusions and Future Work}
\label{sec:conclusion}

In this work, we have shown that reparameterizing the latent space in GANs as a mixture model can lead to a powerful generative model. Via experiments across a diverse set of modalities (digits, hand-drawn object sketches and color photos of objects), we have observed that this seemingly simple modification helps stabilize the model and produce diverse samples even in low data scenarios. Currently, our mixture model setup incorporates some simplifying assumptions (diagonal covariance matrix for each component, equally weighted mixture components) which limit the ability of our model to approximate more complex distributions. These parameters can be incorporated into our learning scheme to better approximate the underlying latent distribution. The source code for training DeLiGAN models and computing modified inception score can be accessed at \url{http://val.cds.iisc.ac.in/deligan/}.

\section{Acknowledgements}

We would like to thank our anonymous reviewers for their suggestions, NVIDIA for their contribution of Tesla K40 GPU, Qualcomm India for their support to Ravi Kiran Sarvadevabhatla via the Qualcomm Innovation Fellowship and Google Research India for their travel grant support.

{\small
\bibliographystyle{ieee}

\begin{thebibliography}{10}\itemsep=-1pt

\bibitem{arulkumaran2016improving}
K.~Arulkumaran, A.~Creswell, and A.~A. Bharath.
\newblock Improving sampling from generative autoencoders with markov chains.
\newblock {\em arXiv preprint arXiv:1610.09296}, 2016.

\bibitem{BengioMDR13}
Y.~Bengio, G.~Mesnil, Y.~Dauphin, and S.~Rifai.
\newblock Better mixing via deep representations.
\newblock In {\em ICML (1)}, volume~28 of {\em JMLR Workshop and Conference
  Proceedings}, pages 552--560, 2013.

\bibitem{burda2015importance}
Y.~Burda, R.~Grosse, and R.~Salakhutdinov.
\newblock Importance weighted autoencoders.
\newblock {\em arXiv preprint arXiv:1509.00519}, 2015.

\bibitem{chen2016infogan}
X.~Chen, Y.~Duan, R.~Houthooft, J.~Schulman, I.~Sutskever, and P.~Abbeel.
\newblock Info{GAN}: Interpretable representation learning by information
  maximizing generative adversarial nets.
\newblock {\em arXiv preprint arXiv:1606.03657}, 2016.

\bibitem{denton2015deep}
E.~L. Denton, S.~Chintala, R.~Fergus, et~al.
\newblock Deep generative image models using a￼ laplacian pyramid of
  adversarial networks.
\newblock In {\em Advances in neural information processing systems}, pages
  1486--1494, 2015.

\bibitem{eitz2012humans}
M.~Eitz, J.~Hays, and M.~Alexa.
\newblock How do humans sketch objects?
\newblock {\em ACM Transactions on Graphics (TOG)}, 31(4):44, 2012.

\bibitem{gershman2012nonparametric}
S.~Gershman, M.~Hoffman, and D.~Blei.
\newblock Nonparametric variational inference.
\newblock {\em arXiv preprint arXiv:1206.4665}, 2012.

\bibitem{goodfellow2014generative}
I.~Goodfellow, J.~Pouget-Abadie, M.~Mirza, B.~Xu, D.~Warde-Farley, S.~Ozair,
  A.~Courville, and Y.~Bengio.
\newblock Generative adversarial nets.
\newblock In {\em Advances in Neural Information Processing Systems}, pages
  2672--2680, 2014.

\bibitem{han2016alternating}
T.~Han, Y.~Lu, S.-C. Zhu, and Y.~N. Wu.
\newblock Alternating back-propagation for generator network.
\newblock {\em arXiv preprint arXiv:1606.08571}, 2016.

\bibitem{jaakkola1998improving}
T.~S. Jaakkola and M.~I. Jordan.
\newblock Improving the mean field approximation via the use of mixture
  distributions.
\newblock In {\em Learning in graphical models}, pages 163--173. Springer,
  1998.

\bibitem{jordan1999introduction}
M.~I. Jordan, Z.~Ghahramani, T.~S. Jaakkola, and L.~K. Saul.
\newblock An introduction to variational methods for graphical models.
\newblock {\em Machine learning}, 37(2):183--233, 1999.

\bibitem{kingma2016improving}
D.~P. Kingma, T.~Salimans, and M.~Welling.
\newblock Improving variational inference with inverse autoregressive flow.
\newblock {\em arXiv preprint arXiv:1606.04934}, 2016.

\bibitem{kingma2013auto}
D.~P. Kingma and M.~Welling.
\newblock Auto-encoding variational bayes.
\newblock {\em arXiv preprint arXiv:1312.6114}, 2013.

\bibitem{krizhevsky2009learning}
A.~Krizhevsky and G.~Hinton.
\newblock Learning multiple layers of features from tiny images.
\newblock 2009.

\bibitem{krizhevsky2012imagenet}
A.~Krizhevsky, I.~Sutskever, and G.~E. Hinton.
\newblock Imagenet classification with deep convolutional neural networks.
\newblock In {\em Advances in neural information processing systems}, pages
  1097--1105, 2012.

\bibitem{mnistlecun}
Y.~Lecun and C.~Cortes.
\newblock {The MNIST database of handwritten digits}.

\bibitem{radford2015unsupervised}
A.~Radford, L.~Metz, and S.~Chintala.
\newblock Unsupervised representation learning with deep convolutional
  generative adversarial networks.
\newblock {\em arXiv preprint arXiv:1511.06434}, 2015.

\bibitem{reed2016learning}
S.~Reed, Z.~Akata, S.~Mohan, S.~Tenka, B.~Schiele, and H.~Lee.
\newblock Learning what and where to draw.
\newblock {\em arXiv preprint arXiv:1610.02454}, 2016.

\bibitem{rezende2015variational}
D.~J. Rezende and S.~Mohamed.
\newblock Variational inference with normalizing flows.
\newblock {\em arXiv preprint arXiv:1505.05770}, 2015.

\bibitem{russakovsky2015imagenet}
O.~Russakovsky, J.~Deng, H.~Su, J.~Krause, S.~Satheesh, S.~Ma, Z.~Huang,
  A.~Karpathy, A.~Khosla, M.~Bernstein, et~al.
\newblock Imagenet large scale visual recognition challenge.
\newblock {\em International Journal of Computer Vision}, 115(3):211--252,
  2015.

\bibitem{salimans2016improved}
T.~Salimans, I.~Goodfellow, W.~Zaremba, V.~Cheung, A.~Radford, and X.~Chen.
\newblock Improved techniques for training {GAN}s.
\newblock {\em arXiv preprint arXiv:1606.03498}, 2016.

\bibitem{szegedy2015going}
C.~Szegedy, W.~Liu, Y.~Jia, P.~Sermanet, S.~Reed, D.~Anguelov, D.~Erhan,
  V.~Vanhoucke, and A.~Rabinovich.
\newblock Going deeper with convolutions.
\newblock In {\em Proceedings of the IEEE Conference on Computer Vision and
  Pattern Recognition}, pages 1--9, 2015.

\bibitem{theis2015note}
L.~Theis, A.~v.~d. Oord, and M.~Bethge.
\newblock A note on the evaluation of generative models.
\newblock {\em arXiv preprint arXiv:1511.01844}, 2015.

\bibitem{zhao2016energy}
J.~Zhao, M.~Mathieu, and Y.~LeCun.
\newblock Energy-based generative adversarial network.
\newblock {\em arXiv preprint arXiv:1609.03126}, 2016.

\bibitem{zhou2014learning}
B.~Zhou, A.~Lapedriza, J.~Xiao, A.~Torralba, and A.~Oliva.
\newblock Learning deep features for scene recognition using places database.
\newblock In {\em Advances in neural information processing systems}, pages
  487--495, 2014.

\end{thebibliography}

}

\end{document}